\documentclass[twocolumn,times,final]{elsarticle}

\usepackage{prletters}
\usepackage{framed,multirow}

\usepackage{amssymb}
\usepackage{latexsym}
\usepackage{subcaption}
\usepackage{amsmath}

\usepackage{url}
\usepackage{xcolor}
\usepackage[table]{xcolor}
\usepackage{latexsym}
\usepackage{amssymb}
\usepackage{amsmath}
\usepackage{amsthm}
\usepackage{booktabs}
\usepackage{enumitem}
\usepackage{graphicx}
\usepackage{color}
\usepackage{svg}
\usepackage{multirow}
\usepackage{algorithm}
\usepackage{algpseudocode}
\usepackage{cleveref}

\newif\ifreview
\reviewtrue   % <-- modalità revisione
\newcommand{\rev}[1]{%
    \ifreview
        {\color{black}#1}%
    \else
        #1%
    \fi
}

\definecolor{lightgray}{gray}{0.9}
\definecolor{lightgreen}{rgb}{0.9,1,0.9} % verde tenue

\newcommand{\gain}[1]{\cellcolor{lightgreen}+#1}
\newcommand{\loss}[1]{\cellcolor{lightgray}#1}
\definecolor{newcolor}{rgb}{.8,.349,.1}

\journal{Pattern Recognition Letters}
\begin{document}

\setcounter{page}{1}

\begin{frontmatter}

\title{Take a Peek: Efficient Encoder Adaptation for Few-Shot Semantic Segmentation via LoRA}

\author[1]{Pasquale \surname{De Marinis}\corref{cor1}} 
\cortext[cor1]{Corresponding author}
\ead{pasquale.demarinis@uniba.it}

\author[1]{Gennaro \surname{Vessio}}

\author[1]{Giovanna \surname{Castellano}}
                
\affiliation[1]{organization={Department of Computer Science, University of Bari Aldo Moro},
                % addressline={Address}, 
                city={Bari}, 
                % postcode={Postal Code}, 
                % state={State},
                country={Italy}}

% \received{1 May 2013}
% \finalform{10 May 2013}
% \accepted{13 May 2013}
% \availableonline{15 May 2013}
% \communicated{S. Sarkar}

\begin{abstract}
Few-shot semantic segmentation (FSS) aims to segment novel classes in query images using only a small annotated support set. While prior research has mainly focused on improving decoders, the encoder’s limited ability to extract meaningful features for unseen classes remains a key bottleneck. In this work, we introduce \textit{Take a Peek} (TaP), a simple yet effective method that enhances encoder adaptability for both FSS and cross-domain FSS \rev{by inducing a lightweight \textit{feature-space shift} conditioned on the support set}. TaP leverages Low-Rank Adaptation to fine-tune the encoder on the support set with minimal computational overhead, enabling fast adaptation to novel classes while mitigating catastrophic forgetting. Our method is model-agnostic and can be seamlessly integrated into existing FSS pipelines. Extensive experiments across multiple benchmarks—including COCO $20^i$, Pascal $5^i$, and cross-domain datasets such as DeepGlobe, ISIC, and Chest X-ray—demonstrate that TaP consistently improves segmentation performance across diverse models and shot settings. Notably, TaP delivers significant gains in complex multi-class scenarios, highlighting its practical effectiveness in realistic settings. A rank sensitivity analysis also shows that strong performance can be achieved even with low-rank adaptations, thereby ensuring computational efficiency.
%Moreover, comparative studies reveal that LoRA outperforms alternative low-rank techniques, such as LoKr and LoHa, in both accuracy and runtime.
By addressing a critical limitation in FSS—the encoder’s generalization to novel classes—TaP paves the way toward more robust, efficient, and generalizable segmentation systems. The code is available at \url{https://github.com/pasqualedem/TakeAPeek}.
\end{abstract}

\begin{keyword}
Few-Shot Learning \sep Semantic Segmentation \sep LoRA \sep Domain Shift \sep Deep Neural Networks.
\end{keyword}

\end{frontmatter}

%\linenumbers

%% main text

\section{Introduction}
\label{sec:intro}

\begin{figure*}[t]
    \centering
    \includegraphics[width=\linewidth]{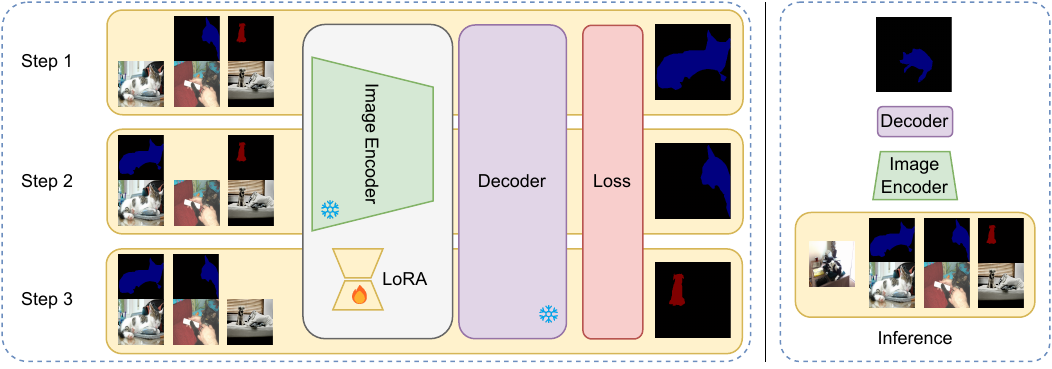}
    \caption{Overview of Take a Peek. Each support image is temporarily treated as a query during adaptation, and its mask supervises a forward-backward pass. Only the LoRA adapter modules are updated. At inference, the adapted encoder processes the query image, while the decoder remains unchanged.}
    \label{fig:TaP}
\end{figure*}

Semantic segmentation has made remarkable progress with large-scale annotated datasets. Nevertheless, this reliance on extensive labeled data limits deployment in real-world scenarios. Models often fail to generalize to novel classes or domains when few or no labeled examples are available.

Few-shot semantic segmentation (FSS) addresses this by segmenting query images using a few annotated supports \cite{shaban2017one}. Early works established prototype-based methods, which rely on global class representations \cite{chenTransformerBasedAdaptivePrototype2024, dong2018few, wang2019panet, zhang2022mfnet}, and affinity-based methods, which compute pixel-wise correlations \cite{chen2024pixel, min2021hypercorrelation, shi2022dense}. While affinity approaches capture fine-grained details, prototype methods are more efficient, especially in multi-class scenarios \cite{marinisLabelAnythingMultiClass2024, zhang2022mfnet}. 
\rev{Recent works also integrated Vision Transformers \cite{zhang2022feature}, Vision-Language Models (VLMs) like CLIP \cite{guoCLIPDrivenPrototypeNetwork2023, liLabelEfficientFewShotSemantic2024}, and backbone fine-tuning strategies such as SVD \cite{sunSingularValueFinetuning}. Related perspectives explore few-shot adaptation of large VLMs using Paramer Efficient Fine Tuning (PEFT) \cite{abd2025optimizing, hu_source-free_2026, madan_revisiting_2023, zeng_harnessing_2026}.}

Despite these advances, most FSS approaches freeze the encoder, adapting only decoders or auxiliary modules. This limits performance: encoders pretrained on ImageNet \cite{heDeepResidualLearning2016a} or Transformers \cite{dosovitskiy2020image} may not produce features aligned with unseen support classes. Even advanced decoders remain constrained. \rev{This is especially problematic in multi-class few-shot settings, where multiple novel categories share a frozen feature space.}

Domain shift introduces additional challenges, addressed by Cross-Domain FSS (CD-FSS) and Test-Time Adaptation (TTA). TTA adapts models to query data at inference. Boudiaf et al.~\cite{boudiafFewShotSegmentationMetaLearning2021} demonstrated transductive fine-tuning on the support set, while others target domain shifts \cite{chenCrossDomainFewShotSemantic2024, leiCrossDomainFewShotSemantic2022}. Most TTA methods adapt decoders or prototypes, keeping the backbone frozen. PEFT and Low-Rank Adaptation (LoRA) have been explored \cite{ bensaidNovelBenchmarkFewShot2025, wangAdaptiveFSSNovel2024}, but efficient encoder adaptation remains underexplored. 
\rev{Related work aligns large VLMs to target concepts with lightweight updates \cite{madan_revisiting_2023}, but those assume foundation models, whereas FSS relies on standard episodic encoders and support-based adaptation.}

\rev{We introduce \textit{Take a Peek} (TaP), allowing the encoder to briefly adapt to the support set at inference. The key idea is to induce a lightweight \textit{feature-space shift} that realigns encoder representations with the current episode. LoRA \cite{huLoRALowRankAdaptation2021} enables efficient, parameter-light updates without modifying backbone weights.}
TaP treats each support image as a temporary query, updating only the encoder via a few gradient steps. It is model-agnostic and integrates with existing FSS pipelines without modifying decoders. Across five SOTA FSS architectures \cite{langLearningWhatNot2022,marinisLabelAnythingMultiClass2024,pengHierarchicalDenseCorrelation2023,shi2022dense,zhang2022feature} and one CD-FSS model \cite{chenCrossDomainFewShotSemantic2024}, TaP consistently improves segmentation on standard and cross-domain benchmarks (Fig.~\ref{fig:TaP}).
\section{Method}
\label{sec:method}

\subsection{Problem Formulation}

In $N$-way $K$-shot semantic segmentation, the goal is to segment objects from $N$ novel classes in a query image, given $K$ annotated examples per class in a support set. The input consists of a support set and a corresponding query image, called an ``episode'', which mimics the few-shot setting during both training and evaluation. Formally, the support set in an episode is defined as $\mathcal{S} = \{(\mathbf{X}_i, \mathbf{Y}_i)\}_{i=1}^{N \times K}$, where $\mathbf{X}_i$ is a support image and $\mathbf{Y}_i$ its ground truth mask. The task within the episode is to segment a query image $\mathbf{X}_q$, which may contain between $0$ and $N$ instances of the target classes defined in $\mathcal{S}$.

\subsection{Take a Peek}
 
We propose Take a Peek, a lightweight and effective fine-tuning strategy that enhances the encoder's adaptability to novel classes. The key idea is to treat the support set as a compact training dataset and use it to briefly adapt the encoder before predicting the segmentation mask for the query image.

Consider a generic encoder-decoder architecture: $\mathcal{E}(\mathbf{X})$ denotes the encoder that extracts features from an input image $\mathbf{X}$, and $\mathcal{D}(\mathbf{Z})$ is the decoder that generates the segmentation mask from the encoded features $\mathbf{Z}$. The encoder is typically pre-trained on a large-scale dataset such as ImageNet \cite{krizhevskyImageNetClassificationDeep2012a}, allowing it to learn general-purpose representations. The decoder, in contrast, is trained on segmentation-specific datasets, such as COCO \cite{linMicrosoftCOCOCommon2015}, usually in an episodic manner, with the encoder frozen. Instead of modifying the decoder, we propose a lightweight adaptation of the encoder using only the support set available during inference. This fine-tuning step improves the encoder's ability to extract features relevant to previously unseen classes while implicitly adapting to the decoder $\mathcal{D}$. Since this adaptation is performed only on the encoder, our method is compatible with any FSS model that follows the encoder-decoder paradigm.

During inference on a generic episode, each support image $\mathbf{X}_i$ with its corresponding mask $\mathbf{Y}_i$ is temporarily treated as a \textit{pseudo-query}. The model attempts to predict $\mathbf{Y}_i$ using the current encoder, guided by a subset of the remaining support images. This approach, which we refer to as \textit{substitution}, allows each support image to serve as a query, enabling the encoder to adapt via supervision from known examples.

At each step, we use a selection strategy $\texttt{Select}$ to choose a subset of the support set $\mathcal{S}_{-i}$ (excluding the current pseudo-query) as contextual support. For most models, we adopt the identity function (i.e., no selection, full support set), except for DMTNet, which is constrained to a 1-shot setting. In that case, we use random sampling to comply with the model's design. The $\texttt{Select}$ function can be customized to reduce computational cost, especially in multi-shot settings where scalability becomes an issue.

The encoder is fine-tuned for $T$ iterations using this episodic structure. Since the adaptation is performed at inference time, $T$ is kept small (e.g., five iterations) to ensure efficiency. Higher values can be used when additional resources are available. Once the encoder is adapted, it is used to encode the actual query image $\mathbf{X}_q$. The decoder then generates the final segmentation mask $\hat{\mathbf{Y}}_q$ using features from the query and the full support set.
This process is illustrated in Fig.~\ref{fig:TaP} \rev{and described in Algorithm~\ref{alg:tap}}.

To achieve efficient and stable adaptation, we integrate LoRA~\cite{huLoRALowRankAdaptation2021}. 
LoRA introduces trainable low-rank matrices into the linear layers of a pre-trained model, allowing for targeted fine-tuning without modifying the full weight matrices. 
Given a weight matrix \( W \in \mathbb{R}^{m \times n} \), LoRA modifies it as:
\begin{equation}
    W' = W + \alpha \cdot A \cdot B,
\end{equation}
where \( A \in \mathbb{R}^{m \times r} \), \( B \in \mathbb{R}^{r \times n} \) are trainable low-rank matrices, \( \alpha \) is a scaling factor, and \( r \ll \min(m, n) \) is the rank of the adaptation.

\rev{The choice of LoRA rank ($r$) and the number of adaptation iterations ($T$) reflects a trade-off between adaptation speed and stability. 
Higher $r$ allows the encoder to adapt more quickly to the support set, improving performance faster, but it also increases the risk of catastrophic forgetting. 
Lower $r$ provides more stable updates but may require more iterations $T$ to achieve similar performance.}

By training only a fraction of the model parameters, LoRA significantly reduces the computational burden and avoids overfitting or catastrophic forgetting, which is crucial in few-shot learning scenarios. In our implementation, LoRA is applied to the attention layers in Transformer-based encoders, which are central for capturing long-range dependencies and contextual information. It is applied to pointwise ($1 \times 1$) convolutional layers for convolutional architectures, which serve as the linear projections in CNN-based models. By updating only these low-rank matrices, we enable efficient adaptation while preserving the generalization capabilities of the pre-trained encoder.

\begin{algorithm}[t]
    \caption{Take a Peek}
    \label{alg:tap}
    %\rev{
    \begin{algorithmic}[1]
    \Require Support set $\mathcal{S} = \{(\mathbf{X}_i, \mathbf{Y}_i)\}_{i=1}^{N \times K}$, query image $\mathbf{X}_q$, encoder $\mathcal{E}$, decoder $\mathcal{D}$, selection strategy $\texttt{Select}$, iterations $T$
    \Ensure Predicted segmentation mask $\hat{\mathbf{Y}}_q$
    
    \State Initialize $\mathcal{E'} \leftarrow \mathcal{E}$ with LoRA modules
    
    \For{$t = 1$ to $T$}
        \For{$i = 1$ to $N \times K$}
            \State $\mathcal{S}_{-i} \leftarrow \{(\mathbf{X}_j, \mathbf{Y}_j)\ |\ j \neq i\}$ \Comment{Exclude $i$}
            \State $\mathcal{C}_i \leftarrow \texttt{Select}(\mathcal{S}_{-i})$ \Comment{Select support subset}
            \State $\{\mathbf{Z}_j\}_{( \mathbf{X}_j, \mathbf{Y}_j ) \in \mathcal{C}_i} \leftarrow \mathcal{E'}(\{\mathbf{X}_j\})$
            \State $\mathbf{Z}_i \leftarrow \mathcal{E'}(\mathbf{X}_i)$
            \State $\hat{\mathbf{Y}}_i \leftarrow \mathcal{D}(\mathbf{Z}_i, \{(\mathbf{Z}_j, \mathbf{Y}_j)\}_{( \mathbf{X}_j, \mathbf{Y}_j ) \in \mathcal{C}_i})$
            \State Compute loss $\mathcal{L}_{FL}(\hat{\mathbf{Y}}_i, \mathbf{Y}_i)$
            \State Update only LoRA parameters in $\mathcal{E'}$ via backprop
        \EndFor
    \EndFor
    
    \State $\mathbf{Z}_q \leftarrow \mathcal{E'}(\mathbf{X}_q)$
    \State $\{\mathbf{Z}_i\}_{i=1}^{N \times K} \leftarrow \mathcal{E'}(\{\mathbf{X}_i\}_{i=1}^{N \times K})$ \Comment{Encode full support set}
    \State $\hat{\mathbf{Y}}_q \leftarrow \mathcal{D}(\mathbf{Z}_q, \{(\mathbf{Z}_i, \mathbf{Y}_i)\}_{i=1}^{N \times K})$ \Comment{No selection at inference}
    \State \Return $\hat{\mathbf{Y}}_q$
    %}
    \end{algorithmic}
\end{algorithm}

During fine-tuning, we optimize the encoder with the Focal Loss~\cite{linFocalLossDense2017}, weighted by the inverse log frequency of class occurrences to address the class imbalance typical of few-shot segmentation. 

TaP improves feature extraction from unseen categories by combining support-based fine-tuning, efficient LoRA adaptation, and a class imbalance-focused loss function, all with minimal computational cost.
\section{Experiments}
\label{sec:experiments}

\begin{table*}[tb]
\caption{
Performance comparison on the COCO $20^i$ dataset and Pascal $5^i$ dataset under 1-way 5-shot and 2-way 5-shot settings.
Absolute mIoU scores are reported in the \textit{Vanilla} rows, and subsequent variants show relative improvements or drops (\(\pm\)) w.r.t.~the corresponding Vanilla baseline.
}
\label{tab:merged}
    \centering
    % Define a smaller column separation value (default is ~6pt)
    \setlength{\tabcolsep}{2pt} 
\resizebox{\textwidth}{!}{
\begin{tabular}{ll|rrrrr|rrrrr|rrrrr|rrrrr}
\toprule
\multirow{3}{*}{Model} &
\multirow{3}{*}{Version} &
\multicolumn{10}{c|}{COCO $20^i$} &
\multicolumn{10}{c}{Pascal $5^i$} \\
\cmidrule(l){3-12} \cmidrule(l){13-22}
& &
\multicolumn{5}{c|}{1-way 5-shots} &
\multicolumn{5}{c|}{2-ways 5-shots} &
\multicolumn{5}{c|}{1-way 5-shots} &
\multicolumn{5}{c}{2-ways 5-shots} \\
\cmidrule(l){3-7} \cmidrule(l){8-12} \cmidrule(l){13-17} \cmidrule(l){18-22}
& &
\multicolumn{1}{c}{Fold 0} &
\multicolumn{1}{c}{Fold 1} &
\multicolumn{1}{c}{Fold 2} &
\multicolumn{1}{c}{Fold 3} &
\multicolumn{1}{c|}{Mean} &
\multicolumn{1}{c}{Fold 0} &
\multicolumn{1}{c}{Fold 1} &
\multicolumn{1}{c}{Fold 2} &
\multicolumn{1}{c}{Fold 3} &
\multicolumn{1}{c|}{Mean} &
\multicolumn{1}{c}{Fold 0} &
\multicolumn{1}{c}{Fold 1} &
\multicolumn{1}{c}{Fold 2} &
\multicolumn{1}{c}{Fold 3} &
\multicolumn{1}{c|}{Mean} &
\multicolumn{1}{c}{Fold 0} &
\multicolumn{1}{c}{Fold 1} &
\multicolumn{1}{c}{Fold 2} &
\multicolumn{1}{c}{Fold 3} &
\multicolumn{1}{c}{Mean} \\
\midrule
\multirow{3}{*}{BAM} &
Vanilla &
44.11 & 52.28 & 48.77 & 46.47 & 47.91 &
35.31 & 41.62 & 44.95 & 40.74 & 40.65 &
67.96 & 71.42 & 65.9 & 61.88 & 66.79 &
40.64 & 48.66 & 55.74 & 52.37 & 49.35 \\
& Decoder FT &
\gain{2.30} & \gain{0.44} & \gain{3.00} & \gain{3.39} & \gain{2.28} &
\gain{2.40} & \gain{4.49} & \gain{0.98} & \gain{1.23} & \gain{2.28} &
\gain{0.40} & \textbf{\gain{0.36}} & \gain{1.90} & \gain{1.86} & \gain{1.13} &
\gain{5.17} & \gain{3.61} & \gain{1.93} & \loss{--0.28} & \gain{2.61} \\
& TaP &
\textbf{\gain{6.78}} & \textbf{\gain{6.62}} & \textbf{\gain{8.00}} & \textbf{\gain{7.18}} & \textbf{\gain{7.14}} &
\textbf{\gain{8.99}} & \textbf{\gain{9.31}} & \textbf{\gain{7.72}} & \textbf{\gain{7.28}} & \textbf{\gain{8.33}} &
\textbf{\gain{2.62}} & \loss{--0.86} & \textbf{\gain{3.67}} & \textbf{\gain{4.01}} & \textbf{\gain{2.36}} &
\textbf{\gain{14.33}} & \textbf{\gain{7.55}} & \textbf{\gain{6.82}} & \textbf{\gain{5.28}} & \textbf{\gain{8.50}} \\
\midrule
\multirow{4}{*}{DCAMA} &
Vanilla &
56.39 & 57.63 & 59.84 & 58.43 & 58.07 &
44.39 & 45.39 & 48.87 & 47.24 & 46.47 &
70.62 & 71.04 & 52.83 & 62.19 & 64.17 &
53.69 & 57.99 & 47.93 & 48.58 & 52.05 \\
& Decoder FT &
\loss{--0.89} & \loss{--1.87} & \loss{--2.00} & \loss{--2.08} & \loss{--1.71} &
\gain{1.88} & \textbf{\gain{3.08}} & \gain{4.97} & \textbf{\gain{4.50}} & \gain{3.61} &
\loss{--0.24} & \loss{--2.59} & \loss{--1.50} & \loss{--1.30} & \loss{--1.41} &
\gain{5.96} & \gain{6.85} & \gain{4.28} & \gain{7.24} & \gain{6.08} \\
& AdaptiveFSS &
\gain{1.78} & \gain{0.62} & \gain{0.56} & \gain{0.93} & \gain{0.97} &
\gain{2.98} & \gain{0.80} & \gain{1.66} & \gain{3.53} & \gain{2.25} &
\textbf{\gain{0.56}} & \loss{--0.03} & \gain{1.01} & \gain{1.19} & \gain{0.68} &
\gain{2.37} & \gain{2.38} & \loss{--0.40} & \gain{3.52} & \gain{1.97} \\
& TaP &
\textbf{\gain{3.44}} & \textbf{\gain{1.30}} & \textbf{\gain{1.10}} & \textbf{\gain{1.10}} & \textbf{\gain{1.74}} &
\textbf{\gain{6.82}} & \gain{2.97} & \textbf{\gain{7.95}} & \gain{4.01} & \textbf{\gain{5.44}} &
\loss{--0.21} & \textbf{\gain{0.76}} & \textbf{\gain{15.25}} & \textbf{\gain{4.55}} & \textbf{\gain{5.09}} &
\textbf{\gain{6.57}} & \textbf{\gain{7.27}} & \textbf{\gain{13.61}} & \textbf{\gain{13.76}} & \textbf{\gain{10.30}} \\
\midrule
\multirow{4}{*}{FPTrans} &
Vanilla &
52.3 & 57.4 & 55.7 & 53.5 & 54.72 &
44.1 & 48.79 & 51.49 & 47.78 & 48.04 &
68.9 & 72 & 74.2 & 65.3 & 70.1 &
61.37 & 67.4 & 74.1 & 62.7 & 66.39 \\
& Decoder FT &
\gain{0.08} & \gain{0.06} & \loss{--0.07} & \loss{--0.09} & \loss{0.00} &
\gain{0.39} & \gain{0.41} & \gain{0.50} & \gain{0.59} & \gain{0.47} &
\gain{0.02} & \gain{0.25} & \gain{0.34} & \loss{--0.02} & \gain{0.15} &
\gain{0.21} & \loss{--0.03} & \loss{--0.04} & \gain{0.16} & \gain{0.08} \\
& AdaptiveFSS &
\loss{--1.95} & \loss{--2.53} & \loss{--3.06} & \loss{--2.15} & \loss{--2.42} &
\loss{--6.66} & \loss{--4.16} & \loss{--4.36} & \loss{--4.52} & \loss{--4.92} &
\loss{--1.48} & \loss{--0.79} & \gain{0.13} & \loss{--1.31} & \loss{--0.86} &
\loss{--6.86} & \loss{--4.64} & \loss{--5.48} & \loss{--3.81} & \loss{--5.20} \\
& TaP &
\textbf{\gain{1.30}} & \textbf{\gain{0.50}} & \textbf{\gain{0.30}} & \textbf{\gain{0.50}} & \textbf{\gain{0.66}} &
\textbf{\gain{3.70}} & \textbf{\gain{4.99}} & \textbf{\gain{3.15}} & \textbf{\gain{4.02}} & \textbf{\gain{3.96}} &
\textbf{\gain{1.30}} & \textbf{\gain{1.70}} & \textbf{\gain{2.40}} & \textbf{\gain{1.30}} & \textbf{\gain{1.68}} &
\textbf{\gain{2.14}} & \textbf{\gain{3.80}} & \textbf{\gain{2.50}} & \textbf{\gain{3.20}} & \textbf{\gain{2.91}} \\
\midrule
\multirow{3}{*}{HDMNet} &
Vanilla &
51.81 & 62.24 & 53.01 & 55.07 & 55.53 &
47.99 & 57.46 & 49.23 & 51.49 & 51.54 &
68.37 & 78.2 & 75.6 & 77.13 & 74.82 &
58.15 & 70.93 & 73.68 & 77.42 & 70.05 \\
& Decoder FT &
\gain{2.02} & \textbf{\gain{1.38}} & \gain{2.37} & \gain{0.73} & \gain{1.63} &
\gain{1.94} & \gain{1.08} & \gain{1.31} & \textbf{\gain{2.17}} & \gain{1.63} &
\textbf{\gain{2.87}} & \textbf{\gain{1.89}} & \gain{4.74} & \textbf{\gain{0.87}} & \textbf{\gain{2.60}} &
\gain{2.79} & \textbf{\gain{3.06}} & \gain{3.88} & \textbf{\gain{1.17}} & \gain{2.72} \\
& TaP &
\textbf{\gain{2.93}} & \loss{--0.84} & \textbf{\gain{3.71}} & \textbf{\gain{0.85}} & \textbf{\gain{1.66}} &
\textbf{\gain{3.21}} & \textbf{\gain{2.75}} & \textbf{\gain{8.23}} & \gain{1.70} & \textbf{\gain{3.97}} &
\gain{1.36} & \loss{--0.96} & \textbf{\gain{5.30}} & \gain{0.01} & \gain{1.43} &
\textbf{\gain{6.78}} & \gain{2.82} & \textbf{\gain{6.44}} & \gain{0.91} & \textbf{\gain{4.23}} \\
\midrule
\multirow{3}{*}{Label Anything} &
Vanilla &
36.12 & 43.32 & 40.35 & 42.87 & 40.66 &
29.71 & 33.46 & 34.13 & 36.65 & 33.49 &
59.93 & 63.83 & 51.00 & 50.78 & 56.39 &
43.40 & 53.50 & 48.80 & 48.60 & 48.58 \\
& Decoder FT &
\gain{4.70} & \textbf{\gain{1.37}} & \gain{0.28} & \gain{0.73} & \gain{1.77} &
\gain{4.59} & \gain{2.19} & \gain{2.50} & \gain{2.38} & \gain{2.91} &
\gain{2.23} & \gain{0.78} & \gain{0.75} & \loss{--0.15} & \gain{0.89} &
\gain{8.42} & \gain{4.48} & \gain{2.23} & \gain{3.00} & \gain{4.53} \\
& TaP &
\textbf{\gain{5.51}} & \gain{0.84} & \textbf{\gain{5.30}} & \textbf{\gain{1.59}} & \textbf{\gain{3.32}} &
\textbf{\gain{5.66}} & \textbf{\gain{4.26}} & \textbf{\gain{6.04}} & \textbf{\gain{4.05}} & \textbf{\gain{5.00}} &
\textbf{\gain{5.95}} & \textbf{\gain{2.55}} & \textbf{\gain{14.79}} & \textbf{\gain{5.50}} & \textbf{\gain{7.19}} &
\textbf{\gain{8.90}} & \textbf{\gain{6.70}} & \textbf{\gain{10.20}} & \textbf{\gain{7.60}} & \textbf{\gain{8.34}} \\
\bottomrule
\end{tabular}
}
\end{table*}

We evaluated our method on widely used benchmark datasets for few-shot semantic segmentation and cross-domain FSS. We set the LoRA rank to $r = 2^{6}$ and fine-tuned the encoder for eight iterations per episode. Performance is reported in terms of mean IoU (\%) across different shot settings. Each method is compared with its vanilla version (without TaP), and results are averaged across 5 runs with 1000 episodes each. We compare our method against a simple baseline that fine-tunes the decoder while keeping the encoder frozen. This baseline is referred to as \textit{Decoder FT} in the tables. Also, we include AdaptiveFSS \cite{wangAdaptiveFSSNovel2024}, a recent method that fine-tunes additional prototypes in the encoder. This method was originally introduced as an additional adaptation stage after meta-training. In our work, however, we modify the procedure to operate at test time. Specifically, AdaptiveFSS was reimplemented and adapted from the authors’ models for test-time use, since the original implementation is not provided as a direct plug-and-play solution.

For the encoder backbones, we used ResNet-50 for HDMNet \cite{pengHierarchicalDenseCorrelation2023}, BAM \cite{langLearningWhatNot2022}, and DMTNet \cite{heDeepResidualLearning2016a}; Swin-B for DCAMA \cite{liuSwinTransformerHierarchical2021}; and ViT-B for Label Anything \cite{dosovitskiyImageWorth16x162020} and FPTrans \cite{zhang2022feature}. The focusing parameter $\gamma$ of the Focal Loss was set to $2$ following recommendations in \cite{linFocalLossDense2017}. We employed the Adam optimizer with a learning rate of $1\text{e}^{-3}$ for DCAMA and Label Anything, and $1\text{e}^{-4}$ for BAM, HDMNet, and DMTNet. The batch size was set to 1, so that weights were updated only on the current episode. All experiments were conducted on an NVIDIA A100 GPU with 64GB of memory.

\subsection{Comparison with the State-of-the-Art}

We report results on the COCO $20^i$ and Pascal $5^i$ datasets in 
%\cref{tab:coco} and \cref{tab:pascal}.
Table~\ref{tab:merged}.
These datasets partition the original classes into four folds, enabling cross-validation on unseen classes. Methods are evaluated in 1-way and 2-way 5-shot scenarios. \rev{ We discuss the 1-shot scenario separately in \cref{sec:1shot} due to the unique challenges it presents.
}

On COCO $20^{i}$, TaP consistently yields the largest improvements across backbones. For instance, BAM gains +7.14\% in the 1-way 5-shot setting and +8.33\% in the 2-way 5-shot setting. DCAMA also shows clear benefits, particularly in the more challenging 2-way case, where TaP improves performance by +5.44\%. In contrast, Decoder FT and AdaptiveFSS produce smaller and less stable gains, and in some cases even lead to performance drops. It is worth noting that AdaptiveFSS was originally designed for a different adaptation scenario, with a tuning set, which may explain its limited effectiveness in our test-time adaptation context.

On Pascal $5^{i}$, the advantage of TaP is again evident. DCAMA achieves a +5.09\% improvement in the 1-way setting and a substantial +10.30\% in the 2-way setting, clearly surpassing the other baselines. FPTrans and HDMNet also benefit consistently from TaP, confirming its robustness across diverse architectures.  

Overall, TaP emerges as the most reliable adaptation strategy, with gains that are particularly pronounced in a 2-way setting, demonstrating its effectiveness under more complex segmentation tasks. However, even being the most reliable, in some cases, it can lead to small performance drops, e.g., in the 1-way 5-shot setting with HDMNet on Fold 1. A workaround could be to evaluate the model on the support set after adaptation and choose the best iteration. One image can be sufficient for this evaluation, as the support set is typically small; we leave this for future work.

\begin{table}[t]  
\caption{Comparison of DMTNet with and without the proposed TaP method on CD-FSS datasets. Results are reported as mean IoU across varying shot counts.}
    \label{table:cd_fss}
    \centering
    \resizebox{\linewidth}{!}{
\begin{tabular}{@{}r|rrrrr@{}}
\toprule
Dataset & Version & 3-shot       & 5-shot       & 10-shot      & 15-shot      \\ \midrule
\multirow{3}{*}{DeepGlobe} & Vanilla    & 49.42                & 51.76                & 54.78                & 54.35                \\
                           & Decoder FT & \gain{0.16}          & \gain{0.27}          & \gain{0.47}          & \gain{0.95}          \\
                           & TaP        & \textbf{\gain{1.64}} & \textbf{\gain{2.42}} & \textbf{\gain{2.83}} & \textbf{\gain{4.55}} \\ \midrule
\multirow{3}{*}{ISIC}      & Vanilla    & 50.01                & 53.77                & 57.13                & 58.89                \\
                           & Decoder FT & \gain{0.69}          & \gain{0.21}          & \gain{0.55}          & \gain{0.81}          \\
                           & TaP        & \textbf{\gain{3.26}} & \textbf{\gain{2.26}} & \textbf{\gain{4.01}} & \textbf{\gain{4.97}} \\ \midrule
\multirow{3}{*}{Chest X-ray}      & Vanilla    & 64.83                & 65.58                & 67.10                & 66.11                \\
                           & Decoder FT & \gain{3.36}          & \gain{5.21}          & \gain{7.96}          & \gain{10.37}         \\
                           & TaP        & \textbf{\gain{13.76}} & \textbf{\gain{15.95}} & \textbf{\gain{18.28}} & \textbf{\gain{20.65}} \\ \bottomrule
\end{tabular}
    }

\end{table}

We further evaluated TaP on DMTNet using three cross-domain benchmarks. DeepGlobe~\cite{demirDeepGlobe2018Challenge2018} covers remote sensing, with seven land cover classes including urban, vegetation, and water. ISIC~\cite{codellaSkinLesionAnalysis2019} focuses on skin lesion segmentation with multiple lesion types. Chest X-ray~\cite{LungSegmentationChest} contains radiographs with normal and pathological lungs annotated at the pixel level.

Across all datasets and shot counts, TaP provides consistent and often substantial improvements. The largest gain is observed on Chest X-ray, where TaP improves mean IoU by up to +20.65\% in the 15-shot setting. In contrast, decoder fine-tuning provides only marginal improvements, never exceeding +10.4\% and, in many cases, falling below +1\%. This confirms that encoder-side adaptation is not only more effective but also more broadly applicable, since decoder design varies widely across models and precludes a truly model-agnostic solution. 

Another important observation is that TaP’s gains scale with the number of shots: improvements are modest in the 3-shot case (e.g., +1.64\% on DeepGlobe) but become pronounced in higher-shot scenarios, suggesting that TaP leverages richer support sets to better adapt encoders to target domains. Moreover, results on DMTNet show that TaP remains effective with random support selection, using only one support image per class for adaptation. This enables efficient adaptation without requiring carefully curated support sets.

\subsection{Rank-iteration Analysis}

\begin{figure}[!t]
    \centering
    \includesvg[width=\linewidth]{figures/rank.svg}
    \caption{Impact of LoRA rank on model performance over adaptation iterations. Results are reported for the DCAMA model on the COCO dataset under the 1-way 5-shot setting.}
    \label{fig:rank}
\end{figure}

We conducted an ablation study to investigate the impact of the LoRA rank together with the number of iterations $T$ on the performance of the DCAMA model. Figure~\ref{fig:rank} illustrates how performance varies across rank values over successive adaptation iterations.
The results show that the rank value plays a significant role in model effectiveness. Initially, performance improves as the rank increases, stabilizing around $r = 2^{8}$. However, performance degrades substantially at $r = 2^{9}$ and $r = 2^{10}$, a trend that persists across iterations and suggests the onset of catastrophic forgetting due to overparameterization.

In contrast, lower ranks exhibit a different behavior: performance tends to peak early and then slightly decline, likely due to underfitting. Higher ranks achieve peak performance in fewer iterations, indicating faster adaptation. However, due to inherent randomness in the few-shot setting and the limited number of iterations, slight oscillations in performance are observed across ranks.

Table \ref{table:rank} summarizes the number of trainable parameters and their proportion relative to the total model size for different ranks. As expected, higher ranks increase the number of trainable parameters, with $r = 2^{10}$ training over 49 million parameters—more than half the total model size. Remarkably, substantial performance improvements are achieved even at the lowest rank ($r = 2^{3}$), where only 0.39 million parameters are trained (just 0.41\% of the total), highlighting the efficiency of our approach.

\begin{table}[tb]
    \centering
    \caption{Number and percentage of trainable parameters as a function of the LoRA rank. Results refer to the DCAMA model with a ResNet-50 backbone.}
    \label{table:rank}
\resizebox{\linewidth}{!}{
    \begin{tabular}{lrrrrrrrrr}
        \toprule
        Rank ($r$) & $2^{3}$ & $2^{4}$ & $2^{5}$ & $2^{6}$ & $2^{7}$ & $2^{8}$ & $2^{9}$ & $2^{10}$ \\
        \midrule
        Parameters (M) & 0.39 & 0.77 & 1.54 & 3.08 & 6.16 & 12.32 & 24.64 & 49.28 \\
        Percentage (\%) & 0.41 & 0.83 & 1.66 & 3.31 & 6.63 & 13.25 & 26.5 & 53.01 \\
        \bottomrule
    \end{tabular}
}
\end{table}

\begin{figure*}[tb]
    \centering
    \includegraphics[width=\linewidth]{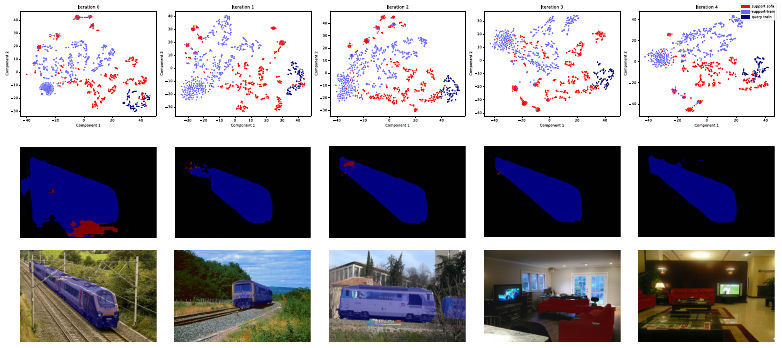}
    \caption{\textbf{Top:} t-SNE visualizations of the query image and support images' class-specific feature embeddings across different stages of the TaP adaptation process. \textbf{Middle:} Segmentation predictions at each adaptation step, showing progressive refinement. \textbf{Bottom:} Input data with overlaid ground truth masks (\textcolor{blue}{blue} for \textit{train}, \textcolor{red}{red} for \textit{sofa}); from left to right: query image, support images for the \textit{train} class (shots 1 and 2), and support images for the \textit{sofa} class (shots 1 and 2).}
    \label{fig:components}
\end{figure*}

\subsection{Qualitative Evaluation}

To better understand how TaP improves class discrimination, we visualize the evolution of feature embeddings and segmentation maps throughout the adaptation process. \rev{This process can be interpreted as a \textit{feature-space shift}}, where the encoder representations are progressively realigned with the support set during adaptation. Figure \ref{fig:components} shows results from a 2-way 5-shot episode using the BAM model with a ResNet-50 backbone on the COCO dataset. The sampled classes are \textit{train} and \textit{sofa}, with two support shots per class shown for clarity. From the query image, 100 pixels belonging to the \textit{train} class are sampled, along with 100 pixels per class from the support images. For visualization, the high-dimensional features are projected to 2D using t-SNE \cite{van2008visualizing}.

The segmentation maps contain many false positives before applying TaP (iteration 0), and the feature embeddings show poor class separation. Notably, embeddings from the query \textit{train} class cluster near the support embeddings of the \textit{sofa} class. Additionally, fragmented mini-clusters appear within the same class, indicating inconsistent representations across regions of the same object.
After the first adaptation step, the model shifts toward false negatives, likely due to the abrupt update of LoRA parameters. By the second step, segmentation quality improves, although some misclassifications remain, particularly within the \textit{sofa} class. These errors gradually diminish, and by the fourth iteration, they are largely resolved. Concurrently, feature embeddings become more compact and better separated between classes. The query embeddings (e.g., dark blue points for \textit{train}) increasingly align with their corresponding support embeddings (e.g., light blue), indicating enhanced intra-class coherence and inter-class discrimination \rev{and illustrating the progressive \textit{feature-space shift} induced by TaP}.

\subsection{Computational Efficiency}

\begin{table}[tb]
\caption{Comparison of different model-agnostic methods for inference-time adaptation, focusing on their memory usage and computational time costs.}
\label{tab:computation}
\resizebox{\linewidth}{!}{
\begin{tabular}{@{}llrrrr@{}}
\toprule
Model &
  Method &
  \multicolumn{1}{c}{\begin{tabular}[c]{@{}c@{}}Mem.\\ Backward/\\ Forward\\ (MB)\end{tabular}} &
  \multicolumn{1}{c}{\begin{tabular}[c]{@{}c@{}}Mem. \\ Opt. \\ Step \\ (MB)\end{tabular}} &
  \multicolumn{1}{c}{\begin{tabular}[c]{@{}c@{}}Time\\ (ms)\end{tabular}} &
  \multicolumn{1}{c}{\begin{tabular}[c]{@{}c@{}}Trainable\\ Params\\ (M)\end{tabular}} \\ \midrule
\multirow{3}{*}{BAM}            & Full        & 5617  & 1106 & 237 & 51.65  \\
                                & Decoder     & 1721  & 368  & 126 & 4.92   \\
                                & TaP         & 5320  & 341  & 253 & 2.29   \\ \midrule
\multirow{4}{*}{DCAMA}          & Full        & 11234 & 2843 & 391 & 92.98  \\
                                & Decoder     & 5772  & 1485 & 185 & 5.07   \\
                                & AdaptiveFSS & 6128  & 1584 & 180 & 0.12   \\
                                & TaP         & 9221  & 1461 & 332 & 3.08   \\ \midrule
\multirow{4}{*}{FPTrans}        & Full        & 9858  & 2038 & 394 & 159.36 \\
                                & Decoder     & 1648  & 682  & 194 & 0.66   \\
                                & AdaptiveFSS & 3735  & 684  & 244 & 0.19   \\
                                & TaP         & 8062  & 726  & 345 & 3.93   \\ \midrule
\multirow{3}{*}{HDMNet}         & Full        & 27251 & 1094 & 757 & 50.9   \\
                                & Decoder     & 12505 & 361  & 297 & 4.16   \\
                                & TaP         & 29568 & 337  & 733 & 2.29   \\ \midrule
\multirow{3}{*}{Label Anything} & Full        & 4619  & 1984 & 235 & 98.9   \\
                                & Decoder     & 791   & 637  & 99  & 11.72  \\
                                & TaP         & 3065  & 494  & 175 & 2.36   \\ \midrule
\multirow{3}{*}{DMTNet}         & Full        & 6460  & 560 & 218  & 28.15   \\
                                & Decoder     & 1673  & 181  & 167  & 2.59  \\
                                & TaP         & 6383  & 161  & 218 & 1.08   \\ \bottomrule
\end{tabular}
}
\end{table}
% TaP introduces additional inference-time cost due to encoder adaptation. However, the number of trainable parameters remains small (e.g., 3.08M for \(r=2^6\) in DCAMA), making the method practical even on large backbones. The adaptation cost scales with \(K \times T\), where \(K\) is the support size and \(T\) the number of adaptation steps, but this cost is incurred only once per support set.
% As shown in Table \ref{tab:computation}, TaP is consistently lighter than full fine-tuning while enabling stronger adaptation than decoder-only updates. Intermediate activations rather than trainable parameters dominate peak memory, explaining differences across architectures (e.g., DCAMA vs.\ BAM). In practice, the efficiency-performance trade-off can be adjusted via the LoRA rank \(r\), the number of iterations \(T\), or the number of support samples per class, allowing TaP to adapt to different computational budgets.

While TaP improves segmentation performance, it introduces additional computational overhead. Specifically, TaP requires training extra parameters, which increases memory consumption and computational load. Nonetheless, the method remains efficient: it updates only a small subset of parameters—e.g., 3.08M for \( r = 2^6 \) in DCAMA—making it viable even in constrained environments.

Training time increases by \( K \times T \) iterations, where \( K \) is the support set size and \( T \) the number of adaptation steps. This cost, however, is incurred only once per support set. In many practical settings—e.g., robotics or edge computing—the support set remains static during deployment, allowing for offline adaptation without affecting real-time performance.

In interactive use cases such as image editing, real-time constraints are relaxed; here, the improved segmentation quality justifies the added adaptation cost. Moreover, TaP allows a performance-efficiency trade-off. As shown in \cref{fig:rank}, higher ranks (e.g., \( r = 2^7 \)) yield substantial gains---e.g., a +3\% improvement with just one iteration---favoring low-latency scenarios. Conversely, lower ranks permit slower, more accurate adaptation when resources permit. The flexibility of our framework also enables us to reduce \( K \) by selecting a smaller number of support images per class, further mitigating computational costs, as in DMTNet.

As shown in \cref{tab:computation}, the computational profile of TaP depends on the underlying architecture. Peak memory usage is influenced not only by the number of trainable parameters but also by intermediate activations and attention maps, which dominate memory consumption in large backbones. For instance, TaP on BAM requires 5320~MB, whereas DCAMA requires 9221~MB despite training a similar number of parameters (2.29M vs.\ 3.08M). This discrepancy arises from DCAMA producing larger attention maps rather than inefficiency in TaP itself. Crucially, across all models, TaP remains substantially lighter than full fine-tuning, while achieving stronger performance than decoder-only updates.  

Decoder-only fine-tuning is consistently the most memory-efficient baseline, but its capacity for adaptation and stability are limited. TaP occupies a middle ground, introducing moderate overhead while enabling richer adaptation and improved segmentation quality. This trade-off can be tuned via the rank parameter $r$ or by reducing the number of support images per class, as in our DMTNet experiments.

\begin{figure}[tb]
    \centering
    \includegraphics[width=\linewidth]{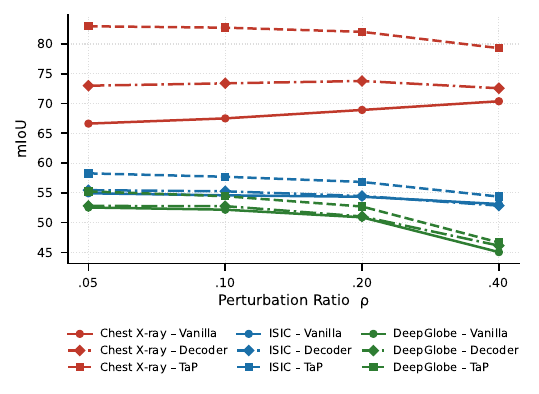}
    \caption{\rev{Performance comparison between TaP and the vanilla version under support set perturbations. The support mask is randomly perturbed by removing parts of it using SLIC superpixels. Results are reported for the DMTNet model on the three cross-domain datasets averaged over 3, 5, 10, and 15 shots.}}
    \label{fig:perturbation}
\end{figure}

\subsection{Robustness to Support Set Perturbations}

\rev{While TaP consistently improves performance across datasets, it may be sensitive to noise in the support set, given its stronger reliance on support mask features. To investigate this, we evaluated the methods on perturbed support sets, where portions of the support mask were randomly removed via SLIC superpixels \cite{achantaSLICSuperpixelsCompared2012}. As shown in Fig.~\ref{fig:perturbation}, TaP exhibits a comparable performance drop to the vanilla baseline across increasing perturbation ratios, suggesting that it does not incur additional sensitivity to support set noise. The same trend is observed for Decoder FT, which follows a similar degradation pattern under increasing perturbations. A notable exception is the Chest X-ray dataset, where the vanilla baseline yields a counterintuitive performance \emph{gain} under perturbation, the reasons for which remain unclear and warrant further investigation.}

\subsection{Stability of TaP}

\rev{As a TTA method, TaP can lead to performance drops in some cases due to the inherent randomness of the few-shot setting. In particular, we observed oscillatory behavior in the early iterations of adaptation, where performance initially degrades before improving, as shown in Fig.~\ref{fig:components}. 
This effect is consistent with the dynamics observed in the qualitative analysis: the first adaptation steps can temporarily misalign query and support features before the feature space progressively shifts toward a support class-conditioned representation. As adaptation proceeds, the model typically converges to a better solution, yielding improved segmentation quality and more coherent feature embeddings.
In some cases, performance may not fully recover within the limited number of iterations, suggesting that further tuning of the adaptation process or the incorporation of additional regularization may be necessary to ensure stability. To maintain the method's simplicity, we used a fixed number of iterations across all episodes, but an adaptive stopping criterion based on validation performance could further mitigate this issue, which we leave for future work.}

\subsection{Discussion on the 1-shot scenario}
\label{sec:1shot}

\begin{table}[t!]
\caption{\rev{1-shot results on COCO-$20^i$ using DCAMA. We report the mIoU (\%) for each fold and the mean value. The green/gray values indicate the performance gain/loss compared to the vanilla baseline.}}
\label{tab:1shot}
\begin{tabular}{@{}rrrrrr@{}}
\toprule
Version & Fold 0 & Fold 1 & Fold 2 & Fold 3 & Mean \\ \midrule
Vanilla     & 50.17       & 53.44       & 52.98        & 51.43       & 52.01       \\
Decoder FT    & \gain{0.26} & \gain{0.44} & \loss{--0.06} & \gain{1.10} & \gain{0.43} \\
AdaptiveFSS & \gain{1.36} & \gain{0.86} & \loss{--0.51} & \gain{0.33} & \gain{0.51} \\
TaP         & \gain{0.34} & \gain{0.22} & \loss{--0.25} & \gain{0.22} & \gain{0.13}  \\ \bottomrule
\end{tabular}
\end{table}

\rev{Test-time adaptation methods for few-shot segmentation (e.g., Take a Peek, decoder fine-tuning, AdaptiveFSS \cite{wangAdaptiveFSSNovel2024}) exploit the support set during evaluation. In the 1-shot case, however, the support set reduces to a single image-mask pair. If this sole image is also used as a pseudo-query, no support data remains, rendering such methods inapplicable.}

\rev{A straightforward workaround is to replicate the single support pair, artificially enlarging the support set. We evaluated this strategy on COCO-$20^i$ using DCAMA as the FSS model \cite{shi2022dense}; the results are reported in Table~\ref{tab:1shot}. Performance was assessed at the iteration yielding the best outcome for each method (iteration 2 for TaP, iteration 3 for decoder fine-tuning, and iteration 4 for AdaptiveFSS). Compared to the 5-shot setting, improvements in the 1-shot case were marginal, highly unstable (appearing only at specific iterations), and occasionally negative.}

\begin{figure}[bt]
    \centering
    \includesvg[width=\linewidth]{figures/results_k1k2.svg}
    \caption{\rev{Performance (mIoU \%) on COCO-$20^i$ using DCAMA in 1-shot and 2-shot settings over different folds and iterations.}}
    \label{fig:results_k1k2}
\end{figure}

\rev{To conduct a deeper analysis, we compared the performance of 1-shot and 2-shot scenarios across all folds and iterations using the DCAMA model for FSS. The results are illustrated in Fig.~\ref{fig:results_k1k2}. The iteration curve for the 2-shot scenario resembles the 5-shot results presented in the main paper, showing an initial increase followed by a period of stabilization or a slight decrease. In contrast, the 1-shot curve has only a slight increase in the first two iterations, followed by a steady decline. This suggests that the model initially benefits from the support pair but subsequently overfits to it, leading to performance degradation.}

\rev{These findings indicate that a single support pair provides insufficient information for meaningful adaptation. Consequently, we excluded the 1-shot scenario from our main experiments. Future research could investigate alternative strategies, such as leveraging data augmentation on the support image to synthetically enrich the support set.}
\section{Conclusion}
\label{sec:conclusion}

In this work, we introduced Take a Peek, a lightweight yet effective method for enhancing the encoder's ability to extract meaningful features from novel classes in both few-shot semantic segmentation and cross-domain FSS. By leveraging Low-Rank Adaptation, TaP enables efficient encoder fine-tuning with minimal computational overhead, enabling rapid adaptation to unseen classes while mitigating the risk of catastrophic forgetting. Importantly, TaP is model-agnostic and can be seamlessly integrated into existing FSS pipelines.

Experiments on standard benchmarks such as COCO \(20^i\) and Pascal \(5^i\), and on CD-FSS datasets including DeepGlobe, ISIC, and Chest X-ray, demonstrate that TaP consistently improves segmentation performance across various architectures, encoders, and shot configurations. The improvements are especially pronounced in more complex scenarios (such as the Chest X-ray dataset), underscoring TaP's effectiveness in challenging settings.
Our rank analysis further revealed that substantial performance gains can be achieved with very low-rank configurations, making TaP a highly efficient solution in terms of both memory and computation. In addition, we profiled the computational cost to confirm the method's efficiency relative to other baselines.
%In addition, we compared LoRA with alternative low-rank adaptation techniques, such as LoKr and LoHa, and found that LoRA strikes the best balance between accuracy and efficiency.

Qualitative analyses confirmed that TaP shifts the feature space to improve class separability and progressively enhances segmentation quality across iterations, reinforcing the value of lightweight encoder adaptation.

TaP addresses a fundamental limitation in FSS by improving encoder adaptability to novel classes, paving the way for more generalizable, accurate, and resource-efficient segmentation models. Future work could explore applying TaP to related tasks such as few-shot object detection or instance segmentation. Another promising direction is to develop a strategy for selecting the optimal adaptation iteration, which could yield larger performance gains while avoiding potential degradation.

\section*{Acknowledgments}

We acknowledge ISCRA for awarding this project access to the LEONARDO supercomputer, owned by the EuroHPC Joint Undertaking, hosted by CINECA (Italy).

\bibliographystyle{model5-names}
\bibliography{main}

\end{document}